\documentclass[accepted]{uai2021}

\usepackage[american]{babel}

\usepackage{hyperref}
\usepackage{url}
\usepackage{xcolor}
\usepackage{times}
\usepackage{amsmath,amsfonts,amsthm,amssymb,mathtools,graphicx,bm,cancel}
\usepackage{subfigure}
\usepackage{comment}
\usepackage{multirow}
\usepackage{multicol}
\usepackage{tabularx}
\usepackage{xr}
\usepackage{natbib} 
\newcommand{\bS}{\bar{S}}

\newcommand\cut[1]{}

\newcommand{\squishlist}{
   \begin{list}{$\bullet$}
    { \setlength{\itemsep}{0pt}      \setlength{\parsep}{3pt}
      \setlength{\topsep}{3pt}       \setlength{\partopsep}{0pt}
      \setlength{\leftmargin}{1.5em} \setlength{\labelwidth}{1em}
      \setlength{\labelsep}{0.5em} } }

\newcommand{\squishlisttwo}{
   \begin{list}{$\bullet$}
    { \setlength{\itemsep}{0pt}    \setlength{\parsep}{0pt}
      \setlength{\topsep}{0pt}     \setlength{\partopsep}{0pt}
      \setlength{\leftmargin}{2em} \setlength{\labelwidth}{1.5em}
      \setlength{\labelsep}{0.5em} } }

\newcommand{\squishend}{
    \end{list}  }

{}
{}
{}
{}

\newcommand{\real}{\mbox{$\mathbb{R}$}}

\newcommand{\rnd}[1]{\left(#1\right)}
\newcommand{\sqr}[1]{\left[#1\right]}

\newcommand{\myvec}[1]{\mbox{$\mathbf{#1}$}}
\newcommand{\myvecsym}[1]{\mbox{$\boldsymbol{#1}$}}

\newcommand{\mymath}[1]{\mbox{$\mathcal{#1}$}}
\newcommand{\bN}{\mbox{$\mymath{N}$}}

\newcommand{\vlambda}{\mbox{$\myvecsym{\lambda}$}}

\newcommand{\vf}{\mbox{$\myvec{f}$}}

\newcommand{\vm}{\mbox{$\myvec{m}$}}

\newcommand{\vx}{\mbox{$\myvec{x}$}}

\newcommand{\vy}{\mbox{$\myvec{y}$}}

\newcommand{\vz}{\mbox{$\myvec{z}$}}

\newcommand{\vF}{\mbox{$\myvec{F}$}}

\newcommand{\vI}{\mbox{$\myvec{I}$}}

\newcommand{\vK}{\mbox{$\myvec{K}$}}

\newcommand{\vU}{\mbox{$\myvec{U}$}}
\newcommand{\vV}{\mbox{$\myvec{V}$}}

\newcommand{\vX}{\mbox{$\myvec{X}$}}

\newcommand{\vZ}{\mbox{$\myvec{Z}$}}

\newcommand{\be}{\begin{equation}}
\newcommand{\ee}{\end{equation}}
\newcommand{\bea}{\begin{eqnarray}}
\newcommand{\eea}{\end{eqnarray}}
\newcommand{\beaa}{\begin{eqnarray*}}
\newcommand{\eeaa}{\end{eqnarray*}}

\title{Subset-of-Data Variational Inference for Deep Gaussian-Processes Regression}

\author[1]{Ayush Jain} 
\author[1]{P. K. Srijith}
\author[2]{Mohammad Emtiyaz Khan}
\affil[1]{%
    Department of Computer Science and Engineering \\
    Indian Institute of Technology Hyderabad, India
}
\affil[2]{%
    RIKEN Center for AI Project \\
    Tokyo, Japan
}

\begin{document}
\maketitle

\begin{abstract}
   Deep Gaussian Processes (DGPs) are multi-layer, flexible extensions of Gaussian Processes but their training remains challenging.~Sparse approximations simplify the training but often require optimization over a large number of inducing inputs and their locations across layers.~In this paper, we simplify the training by setting the locations to a \emph{fixed} subset of data and sampling the inducing inputs from a variational distribution.~This reduces the trainable parameters and computation cost without significant performance degradations, as demonstrated by our empirical results on regression problems.~Our modifications simplify and stabilize DGP training while making it amenable to sampling schemes for setting the inducing inputs. 

\end{abstract}

\section{INTRODUCTION}
Deep Gaussian Processes (DGPs) aim to extend the functional-learning capabilities of Gaussian Processes (GPs) to improve their flexibility. A DGP consists of multiple layers of GPs stacked one over the other~\citep{damianou13, damianou:thesis15} which enables us to model complex functions, such as non-stationary and discontinuous functions. 
DGPs are motivated by the deep architectures used in deep learning but they promise to overcome the limitations of deep learning, e.g., they can deal with smaller datasets, improve uncertainty estimates, and perform model selection.

\begin{figure*}[t]
    \includegraphics[width=5.2cm]{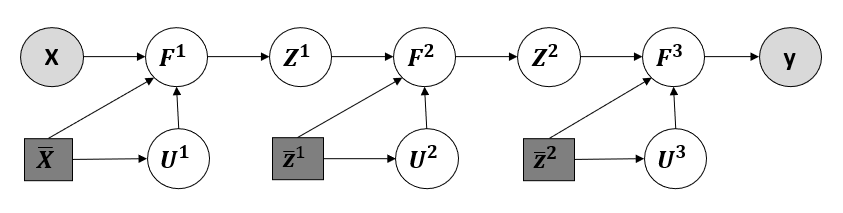}
    \hfill
    \includegraphics[width=5.2cm]{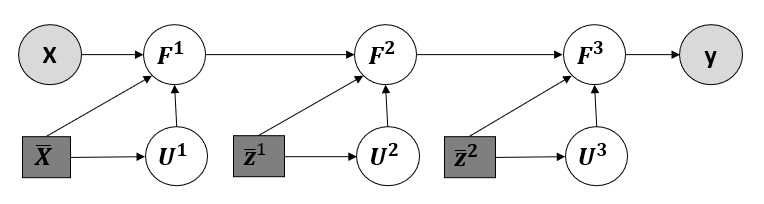}
    \hfill
    \includegraphics[width=5.7cm]{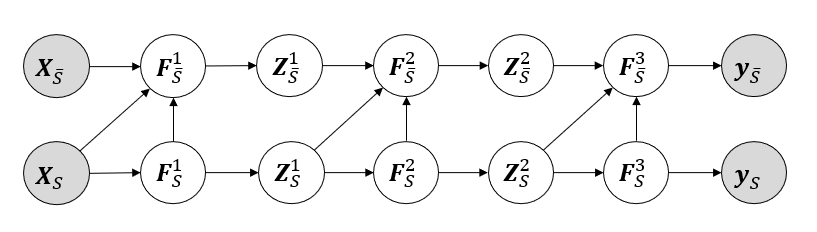}
    \subfigure[\citet{damianou13}]{\includegraphics[width=5.2cm]{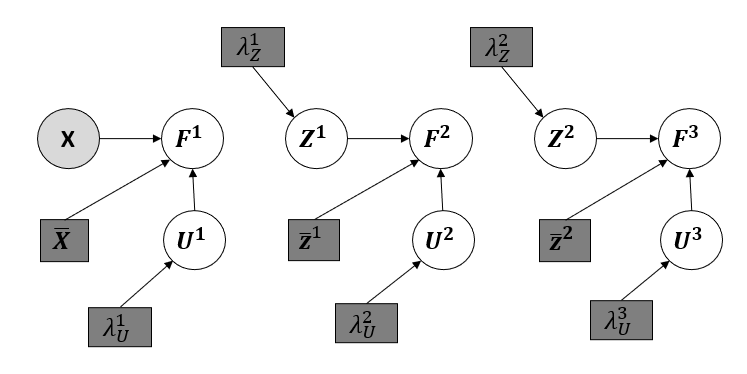} \label{fig:originalDGP}}
    \hfill
    \subfigure[\citet{dsvi}]{\includegraphics[width=5.2cm]{dsviDGP_Feb21.png} \label{fig:dsvi}}
    \hfill
    \subfigure[Our Subset-of-Data Method]{\includegraphics[width=5.7cm]{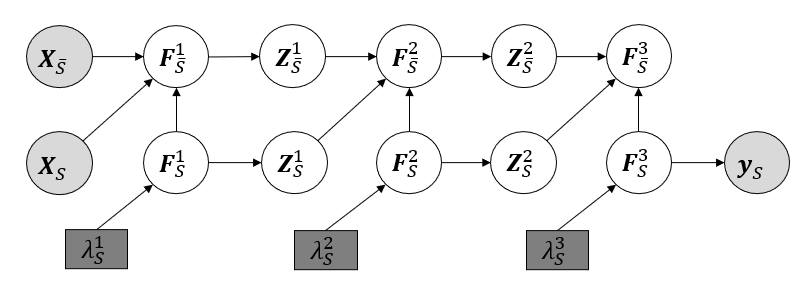} \label{fig:ours}}
  \caption{Top row shows the (augmented) DGP model for input $\vX\in\real^{N\times H}$ and output $\vy \in \real^N$ for three methods, and bottom row shows the corresponding approximate posterior. Trainable parameters are shown with square-gray boxes in the bottom row, and we can see that our subset-of-data method requires the lowest number of parameters. Methods shown in the top row of (a) and (b) both use an augmented DGP model with inducing inputs $\bar{\vZ}^{l} \in \real^{M\times D^l}$ and function values $\vU^l \in \real^{M\times D^l}$. As a consequence, they require estimation of variational parameters $\vlambda_U^l$ for the variational distribution $q(\vU^l)$ and inducing inputs $\bar{\vZ}^l$, as shown in the bottom row. The method in (a) additionally requires variational parameters for $\vZ^l$. Our method shown in (c) replaces $\vU^l$ by the functions $\vF_S^l \in \real^{M\times D^l}$ defined over a subset $S$ of $M$ data examples (see the top row; $\bar{S}$ denotes the set of training examples other than those in $S$). As shown in the bottom row, we only need to learn the parameters $\vlambda_S^l$ of $q(\vF_S^l)$, since the rest of the quantities follow the same relationship as the DGP model. This leads to a reduction in the number of trainable parameters. Additionally, the inputs $\vZ^l$ are not learned but sampled given $\vF^l$.}
  \label{fig:all_methods}
\end{figure*}

Unfortunately, obtaining good performance with DGP on large problems remains a challenge, mainly due the inefficient training procedures currently used. Unlike deep learning, the training procedures for DGPs are computationally expensive, slow to run, and rarely beat the performance of deep learning. Posterior inference is even more challenging than GPs and involve multiple large matrix inversions of size $O(N^3)$ where $N$ is the number of data examples. Existing methods mitigate such difficulties by using \emph{sparse variational inference} (VI)~\citep{titsias2009variational} which augment the model with auxiliary variables called \emph{inducing inputs}.
An example is shown in Fig. \ref{fig:originalDGP} (top row) where inducing input $\bar{\vZ}^l$ and corresponding functions $\vU^l$ are introduced in every layer $l$. These auxiliary quantities are different from the functions and inputs defined over input $\vX$ in the training data (denoted by $\vF^l$ and $\vZ^l$ respectively).
During inference, all the auxiliary quantities need to be estimated along with the variational distribution parameters and GP kernel hyper-parameters to obtain a posterior approximation. Due to this, the number of trainable parameters is often very large.

Various strategies for training have been proposed in the past.
The approach of \cite{damianou13} (Fig. \ref{fig:originalDGP}) assumes a mean-field variational approximation over $\vZ^l$ and $\vU^l$, and estimates their variational parameters (denoted by $\vlambda_Z^l$ and $\vlambda_U^l$ respectively), as well as the inducing inputs $\bar{\vZ}^l$. All these trainable parameters are shown with square gray boxes in the figure.
The doubly stochastic variational inference (DSVI) approach of \cite{dsvi}, shown in Fig. \ref{fig:dsvi}, reduces the number of trainable parameters by integrating $\vZ^l$ out. The approach still needs to estimate the variational parameters  $\vlambda_U^l$ and inducing inputs $\bar{\vZ}^l$.
Our goal in this paper is to further reduce the number of trainable parameters.

Our key idea to reduce the number of parameters is to use a fixed subset of data as inducing inputs, i.e., a subset $S \subset \{1,2,\ldots, N\}$ of size $M$ with $M\ll N$. 
Instead of using auxiliary inputs $\bar{\vZ}^l$ and functions $\vU^l$, we use the inputs $\vZ_S^l$ and functions $\vF_S^l$ defined over the subset $S$; see the top row in Fig. \ref{fig:ours}.~We introduce variational distributions over $\vF_S^l$ with parameters denoted by $\vlambda_S^l$ while keeping the relationship between the rest of the quantities as defined by the DGP model (see the bottom row in Fig. \ref{fig:ours}).  The trainable parameters are $\vlambda_S^l, \forall l$, the size of which is strictly lower than the methods in Fig. \ref{fig:originalDGP} and \ref{fig:dsvi} (assuming the number of inducing inputs $M$ to be the same). 

Our approach exploits the labels $\vy_S$ associated with the inducing input $\vX_S$ to form  the variational posterior and consequently simplifying the computation of the evidence lower-bound. The resulting bound naturally combines the predictive likelihood and marginal likelihood together leading to a simple yet effective solution (see \eqref{sod_vlb} in Section \ref{sect:sod_vi}).  The idea of choosing inducing inputs from training set simplifies the inference due to the associated labels, ultimately leading to a reduction in the number of variational parameters.
The empirical results suggest that our method reduces the cost of DGP inference without a significant degradation in the performance. 
An additional advantage of our method is that the inducing inputs $\vZ_S^l$ are not trained but sampled using the variational distribution over $\vF_S^l$.

In our experiments, we fix the subset $S$ by clustering the data before training, but our methods is amenable to sampling approaches, such as leverage score \citep{alaoui2015fast} and determinantal point process \citep{kathuria2016batched}, where the subsets can be sampled during training. We expect our approach to further improve when augmented with such sampling approaches.
Throughout the paper, we focus primarily on improving the methods of \cite{damianou13} and \cite{dsvi}, but our approach could potentially be useful for many other variants, e.g., methods using amortized inference~\citep{Dai:VAEDGP16}, nested inference~\citep{hensman2014nested}, approximate Expectation Propagation~\citep{bui2016deep}, random Fourier features expansion~\citep{randombonila} and implicit posterior variational inference~\citep{yu19}. 
We only address regression problems, but extension to classification and multi-output DGPs can be obtained by using standard non-conjugate inference procedures.

\section{DEEP GAUSSIAN PROCESSES }

In this section we provide a background on deep Gaussian process models and their training methods.
We consider the regression problem with $N$ training data points, $\vX = \{\vx_n\}_{n=1}^N$ and the corresponding labels ${\vy} = \{y_n\}_{n=1}^N$, where $\vx_n \in \mathcal{R}^{H}$ and $y_n \in \mathcal{R}$. We assume that there exists a regression function $f : \mathcal{R}^{H} \rightarrow \mathcal{R}$ which maps the training data to outputs, and our goal is to learn the function. 

\begin{figure}[t]
  \centering
  \includegraphics[width=8.2cm]{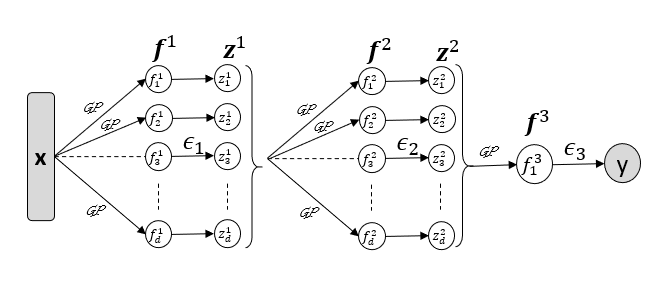}
  \caption{Deep Gaussian Process Model models a scalar output $y$ given an input vector $\vx$ with nested layers of function $\vf^l$, generated using GPs, and their noise versions $\vz^l$.}
  \label{originalDGP}
\end{figure}

Deep Gaussian Processes (DGPs) provide a rich and flexible prior to model functions by stacking several GP priors. The DGP model described in~\cite{damianou13,damianou:thesis15} (shown in Fig.~\ref{originalDGP}) models the regression function by as a composition of several layers of functions, $\vy =  \vf^L \circ (\vf^{L-1} \ldots \circ (\vf^1 (\vX))) $(assuming $L$ layers). The $l^{th}$ layer consists of $D^l$ functions $f^l = \{ f^l_d\}_{d=1}^{D^l}$ mapping representations in layer $l-1$ to obtain $D^l$ dimensional representation for layer $l$. In each layer independent GP priors are placed over the functions $f^l_d$  conditioned on the output of the previous layer, 
\begin{equation}
   f_{d}^{l}(\cdot)  \sim {\mathcal{GP}}( \mu_{d}^{l}(\cdot) , k^{l}(\cdot,\cdot)), 
\end{equation}
where $\mu_{d}^{l}: \mathcal{R}^{D^{l-1}} \rightarrow \mathcal{R}$ is the mean function and $k^{l}: \mathcal{R}^{D^{l-1}} \times \mathcal{R}^{D^{l-1}} \rightarrow \mathcal{R}$ is the covariance function. The functions $f^1_d(\cdot)$ in the first layer act on the inputs $\vx_n$ to produce the mapping $F^1_{n,d} := f^1_d(\vx_n)$. The first-layer representations $Z^1_n$ are obtained by adding noise to these mappings, $Z^{1}_{n,d} = F^1_{n,d} + \epsilon_1$, where $\epsilon_1 \sim \bN(0,\sigma_1^2)$. These are then fed as inputs to the next layer and the process is repeated.
We use $\vZ^l, \vF^l\in\real^{N\times D^l}$ to denote the matrices obtained with entries $Z_{n,d}^l$ and $F_{n,d}^l$ respectively. 

The joint distribution of $\vy$ and $\vF^l,\vZ^l$ over all layers is 
\begin{align}
    &p({\vy},\vF^L, \vZ^{L-1}, \ldots, \vF^2, \vZ^1, \vF^1|\vX)\\
    &=\underbrace{\prod_{n=1}^N p(y_n|F^L_n)}_{\textrm{Likelihood}}
    \underbrace{\left[\prod_{l=1}^{L-1} p(\vF^{l+1}|\vZ^l) p(\vZ^l|\vF^l)\right]p(\vF^1|\vX)}_{\textrm{Deep GP Prior}} \nonumber
\end{align}
Here, the likelihood over the observation $p(y_n|F^L_n) = \bN(y_n; F^L_n, \sigma_{L}^2)$, and the factors associated with intermediate layers can be factorized as 
\begin{align}
p( \vZ^l|\vF^l) &= \prod_{n=1}^N \prod_{d=1}^{D^l} \bN( Z^l_{n,d};F^l_{n,d}, \sigma_l^2)
\end{align}
The prior over the functions at some layer $l$ and dimension $d$ is considered to be a zero mean GP with kernel $k^l(\cdot, \cdot)$, i.e. $f^l(\cdot) = GP(0, k^l(\cdot, \cdot))$. Consequently, the function values over the data points are distributed as a zero mean Gaussian.   
\begin{align}
p({F^l_{:,d}}|\vZ^{l-1}) &= \bN(0,\vK^l_{Z^{l-1},Z^{l-1}}) \nonumber
\end{align}
Here, $\vK^l_{Z^{l-1},Z^{l-1}}$ is an $N \times N$ co-variance matrix  obtained by evaluating the kernel $k^l(\cdot, \cdot)$ on the $l-1$ layer latent representations of all the data points. The priors and likelihood are conditioned on kernel hyper-parameters and noise variance($\sigma_{l}$). Squared exponential kernel is most widely used kernel in this setting which have length scales and variance as hyperparameters. Hyper-parameter learning requires optimisation of marginal likelihood which is intractable in Deep GPs. This is because latent representation appear nonlinearly in the full likelihood due to the form of $p(\vF^{l+1}|\vZ^l)$ terms. Due to underlying intractability, approximate Bayesian methods are used to compute an approximate inference for Deep GPs.

Training Deep GPs require approximate inference methods which largely relies on sparse variational-inference methods ~\citep{damianou13,dsvi}. These methods compute a tractable lower bound for marginal likelihood using sparse GPs~\citep{titsias2009variational,GPbigdata}. These approaches simultaneously addresses intractability and scalability issues in Deep GPs by introducing inducing points with inducing input $\bar{\vZ}^l$ and output $\vU^l$ for each layer $l$, all of which are learnt from the variational lower bound (top figure in Fig.~\ref{fig:originalDGP}).

The posterior inference is simplified by estimating $\bar{\vZ}^l$ and introducing a Gaussian approximation $q(\vU^l)$ with variational parameters $\vlambda_U^l$. Given these two quantities the cost of posterior inference reduces drastically.
The corresponding approximate posterior introduced in ~\cite{damianou13}  is shown at the bottom figure in Fig. \ref{fig:originalDGP} where an additional variational distribution over $\vZ^l$ is used.

Even though the inference is simplified with the method of \cite{damianou13}, the number of variational parameters is still quite large. 
\cite{dsvi} reduce the number by marginalizing $\vZ^l$ thereby not requiring the additional distribution over it. During inference, $\vZ^l$ is obtained by using a forward sampling from the DGP model.
The approach \cite{dsvi} reduces the cost of inference by reducing the number of trainable parameters.

\section{SUBSET-OF-DATA Variational Inference (SOD-VI)}

\label{sect:sod_vi}

Our goal in this paper is to reduce the number of trainable parameters. We view the inducing inputs $\vU^l$ as inputs without any real inputs $\vx$ or labels $y$. Our key idea then is to replace them by latent functions $\vF_S^l$ defined over a subset $S$ of the training data. This reduces the number of trainable parameters, because the input locations $\bar{\vZ}^l$ are replaced by $\vZ_S^l$ which can be sampled from the DGP model, instead of being learned.

We first describe this for a single-layer GP.

\subsection{Subset-of-Data VI for GP}
We first consider deriving the subset-of-data VI for a simple one layer GP model with $\vF$ representing function values at data points $\vX$. We write the  joint likelihood over observations (outputs) and latent function values as 
\begin{align}
p({\vy},\vF) &= \sqr{ \prod_{n=1}^N p(y_n|F_{n})} p(\vF) \nonumber \\
&=  p(\vy_S|\vF_S) p(\vy_{\bar{S}}|\vF_{\bar{S}}) p(\vF_{\bar{S}}|\vF_S)p(\vF_S)
\label{SOD_GPlik}
\end{align}
where $S\subset \{1,2,\ldots,N\}$ is a subset of data examples, $\vF_S$ denotes the latent function values over the inputs indexed by $S$ and $\vy_S$ are the corresponding observations. $\bar{S}$ is the complement of subset $S$ (indexes of data examples not in $S$), and $\vF_{\bar{S}}$ and $\vy_{\bS}$ denote the corresponding latent function values and observations, respectively. The subset $S$ can be chosen randomly, by clustering the training data and choosing data points closest to the centroids, or  by sampling  approaches  such  as  leverage score and determinantal point processes.
Throughout, we use the clustering approach. 
\begin{figure}[t]
  \centering
  \includegraphics[width=2.5in]{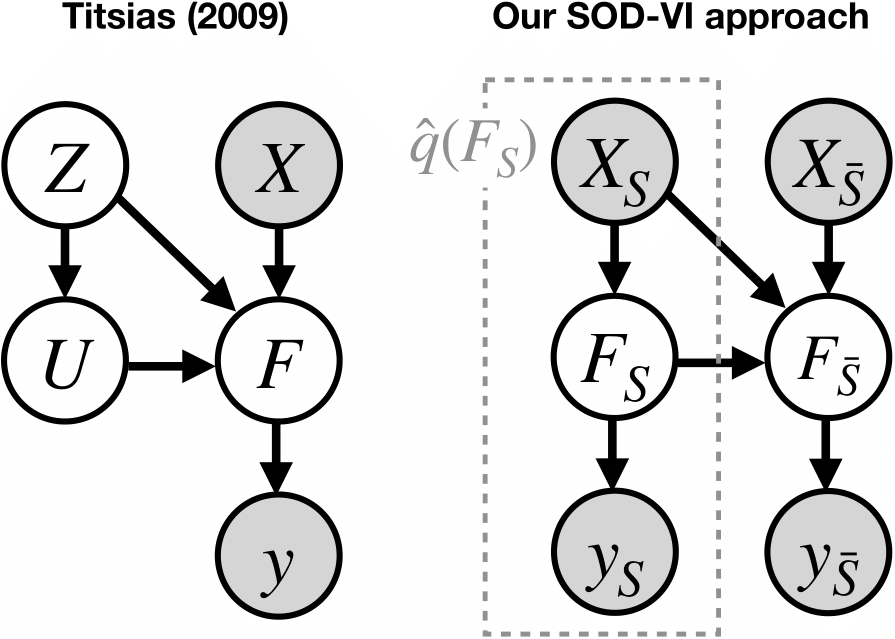}
  \caption{\citet{titsias2009variational}'s approach uses unknown input locations $\vZ$ to define inducing inputs $\vU$. In contrast, we use \emph{known} locations $\vX_S$ from the training data. Since the labels $\vy_S$ for these locations are available, the posterior over $\vF_S$ is also known and fixed (see \eqref{SOD_GPVarPost} for an expression). Our lower bound however has the same form as the one traditionally used in sparse variational GP methods (see \eqref{sod_vlb}).}
  \label{titsias}
\end{figure}

Our goal is to obtain a posterior over $\vF = \{ \vF_{\bar{S}}, \vF_S\}$, while using $\vF_S$ as the inducing points and the corresponding $\vX_S$ as the inducing inputs. To build a variational approximation, we first rewrite the posterior over $\vF$ as follows,

\begin{align}
\label{eqn:sod_pos}
p(\vF|\vy) 
\, \propto \, p(\vy_{\bar{S}}|\vF_{\bar{S}}) p(\vF_{\bar{S}}|\vF_S) \underbrace{ \frac{ p(\vy_S|\vF_S) p(\vF_S)}{p(\vy_S)} }_{= p(\text{\vF}_S|\text{\vy}_S)} 
\end{align}
where we have explicitly grouped the terms corresponding to the posterior $p(\vF_S|\vy_S)$.
We define a variational distribution which has the same structure but use a variational approximation $q(\vF_S)$ instead of the prior $p(\vF_S)$,

\begin{align}
           p(\vF|\vy) \approx \frac{1}{\hat{Z}} p(\vF_{\bar{S}}|\vF_S) \underbrace{ \frac{ p(\vy_S|\vF_S) q(\vF_S)}{p(\vy_S)} }_{= \hat{q}(\text{\vF}_S)},  \label{SOD_GPVarPost}
\end{align}
where $\hat{Z}$ is the normalizing constant.
The distribution $\hat{q}(\vF_S)$ is an approximation to the posterior $p(\vF_S|\vy_S)$, obtained via $q(\vF_S)$. The corresponding posterior over $\vF_{\bS}$ can be obtained by marginalizing over $\vF_S$, which we denote by $q(\vF_{\bS}) = \int p(\vF_{\bS}|\vF_S)\hat{q}(\vF_S) d\vF_S$.

The above approach is different from the standard variational approach of~\citet{titsias2009variational} where the inducing inputs do not have any labels associated with them, and we are forced to use $p(\vF|\vy) \approx p(\vF_{\bS}|\vF_S) \hat{q}(\vF_S)/\hat{Z}$ with an arbitrary free-form Gaussian $\hat{q}(\vF_S)$. In our case, due to the associated label, $\hat{q}(\vF_S) = p(\vF_S|\vy_S)$ which is obtained by choosing $q(\vF_S) = p(\vF_S)$. Our approach results in a simpler variational posterior than the approach of \citet{titsias2009variational}. See Fig. \ref{titsias} for an illustration.

We will now show that, for the defined $\hat{q}(\vF_S)$ and $q(\vF_{\bS} )$, the variational lower bound can be derived as 

\begin{align}
&  E_{p(\text{\vF}_{\bar{S}}|\text{\vF}_S)\hat{q}(\text{\vF}_S)}\sqr{ \log\rnd{ \frac{p(\vy|\vF) p(\vF_{\bar{S}}|\vF_S)p(\vF_S)}{p(\vF_{\bar{S}}|\vF_S)\hat{q}(\vF_S)} }}\nonumber \\
&= E_{q(\mathbf{F}_{\bar{S}})}[\log(p(\vy_{\bar{S}}|\vF_{\bar{S}})]  + E_{\hat{q}(\mathbf{F}_S)}[\log(p(\vy_S|\vF_S)]  \nonumber \\
&\quad\quad\quad\quad\quad\quad\quad\quad\quad\quad- KL(\hat{q}(\vF_S)\|p(\vF_S)).  
\label{sod_vlb}
\end{align}
This is similar to the lower bounds used in \citet{titsias2009variational} and \citet{GPbigdata} but with $\vF_S$ as the inducing points and the distribution set to $\hat{q}(\vF_S)$.
In our case, the distribution exploits the structure shown in \eqref{SOD_GPVarPost} by using the labels, but for \citet{titsias2009variational} this can be an arbitrary Gaussian.
The advantage with our approach is that we do not need to estimate the inducing inputs since they are fixed at the inputs $\vX_S$. Note that in general our approach will give different results than \citep{titsias2009variational} due to the choice of the inducing inputs as well as the variational approximation. We will show in the next section that our approach is useful in reducing number of trainable parameters for Deep GPs.

The variational lower bound \eqref{sod_vlb} takes an interesting form where the first term is the negative log-predictive probability (NLPP) \citep{shevade09} on a validation set $\bar{S}$, and the rest of the terms only depend on $S$. 
Therefore, we can see the lower bound maximization as maximizing the fit over the data both in $S$ and $\bar{S}$, which is a better approach than the classical NLPP based approaches considering $\bar{S}$ alone. The variational parameters $\vlambda_S$ associated with $q(\vF_S)$ (for instance, mean and Covariance parameters associated with a variational  Gaussian approximation) and  hyper-parameters such as kernel parameters are learnt by maximizing the variational lower bound.

\subsection{Subset-of-Data VI for DGP}
We will now extend the subset-of-data approach to DGPs. Fig.~\ref{fig:ours} (top row) shows the graphical model of the DGP model, where we have split the variables into two sets corresponding to examples in $S$ and $\bar{S}$. Comparing to the other methods in Fig. \ref{fig:all_methods}, in our approach the subset $S$ can be seen as playing the role of the inducing points. Consequently, the joint distribution also can be written in a conditional form where the term over $S$ are conditioned on the terms over $\bar{S}$ as shown below:
\begin{align}
&\textrm{Likelihood: } p(\vy_S|\vF_S^L)p(\vy_{\bar{S}}|\vF_{\bar{S}}^L)  \label{trueLik}\\
&\textrm{DGP Prior:}\left[\prod_{l=L-1}^{1} p(\vF_{\bar{S}}^{l+1}|\vF_S^{l+1},\vZ^l)p(\vF_S^{l+1}|\vZ_S^l) p(\vZ^l|\vF^l)\right]\nonumber \\
&\quad\quad\times p(\vF_{\bar{S}}^1|\vF_S^1,\vX)p(\vF_S^1|\vX_S)\label{trueJointPrior}
\end{align}

Similarly to the single-layer GP case, we can simply replace the priors $p(\vF_S^{l+1}|\vZ_S^l)$ by $q(\vF_S^{l+1})$ and keep the rest of the structure same as the original model, to get the following variational approximation: 
\begin{align}
&\bar{q}(\vF^{1:L},\vZ^{1:L-1}) \propto  \nonumber\\ 
&p(\vy_S|\vF_S^L)\left(\prod_{l=L-1}^{1}p(\vF_{\bar{S}}^{l+1}|\vF_S^{l+1},\vZ^{l})q(\vF_S^{l+1})p(\vZ^{l}|\vF^{l})\right)\nonumber\\
&\quad\quad\quad\quad \times p(\vF_{\bar{S}}^{1}|\vF_S^{1},X)q(\vF_S^{1}). \label{qPosterior}
\end{align}

where where $q(\vF_S^l)$ represents the variational distribution over $\vF_S^l$, which is typically a Gaussian with mean and Covariance as variational parameters, $\vlambda_S^l = \{{\boldsymbol{\mu}}_{S}^l,{\boldsymbol{\Sigma}}_{S}^l\}$.
The marginals and conditions derived from the above distributions will be denoted by $q(\cdot)$ to simplify the notation, e.g., $q(\vZ_S^l)$ will be the marginals of $\vZ_S^l$ obtained according to the graphical model shown in Fig. \ref{fig:ours} (bottom row).

There are two main points to note. First, in the graph structure of the posterior approximation shown in Fig. \ref{fig:ours} (bottom row), we see that the directed arrow from $\vZ^l$ to $\vF^l$ is removed.
Second, the inducing inputs $\vZ_S^l$ can be obtained directly from the samples $\vF_S^l$ of the variational distribution. Unlike previous approaches, we do not need to learn them. This is the main reason behind the reduction in the number of trainable parameters with our approach.

Given the posterior approximation, the variational lower bound is directly obtained by using~\eqref{trueLik},~\eqref{trueJointPrior} and~\eqref{qPosterior},  
\begin{align}
&\,\, L= E_{\bar{q}\left( \text{\vF}^{1:L},\text{\vZ}^{1:L-1}\right)} \log\left(\frac{p(\vy,\vF^{1:L},\vZ^{1:L-1}|\vX)}{\bar{q}\left( \text{\vF}^{1:L},\text{\vZ}^{1:L-1}\right) }\right) \nonumber \\
& = E_{q(\mathbf{F}_{\bar{S}}^L)}[\log(p(\vy_{\bar{S}}|\vF_{\bar{S}}^L))] +E_{\hat{q}(\mathbf{F}_{S}^L)}[\log(p(\vy_S|\vF_{S}^L))]\nonumber\\
&\quad- \sum_{d=1}^{D^1} KL(q(\vF_{S,d}^1) \|p(\vF_{S,d}^1|\vX_S))\nonumber\\
&\quad- \sum_{l=2}^{L-1} \sum_{d=1}^{D^l} E_{q(\mathbf{Z}_S^{l-1})}\left[ KL(q(\vF_{S,d}^l) \|p(\vF_{S,d}^l | \vZ_S^{l-1}))\right]\nonumber\\
&\quad-  E_{q(\mathbf{Z}_S^{L-1})}[KL(\hat{q}(\vF_{S}^L)\|p(\vF_{S}^L |\vZ_S^{L-1}))].
\label{SOD_Elbo}
\end{align}
The form of the lower bound is slightly more complicated than \eqref{sod_vlb}, but the last three lines here are simply an expansion of the KL term splitting over the layers and dimensions. Another minor difference is that expectation with respect to $q(\vZ_S^l)$ needs to be taken in the third line. We also note that the derivation can be easily extended to the multi-output case, where the first two terms and the last term  in \eqref{SOD_Elbo} will have an additional summation over the multiple outputs. 

We now derive quantities required to compute the lower bound. We assume a Gaussian variational approximation $q(\vF_{S}^L) =  \mathcal{N}(\vF_{S}^L;{\boldsymbol{\mu}}_{S}^L,{\boldsymbol{\Sigma}}_{S}^L)$ for the last layer and $q(\vF_{S,d}^{l-1}) :=  \mathcal{N}(\vF_{S,d}^{l-1};{\boldsymbol{\mu}}_{S,d}^{l-1},{\boldsymbol{\Sigma}}_{S,d}^{l-1})$ for the layer $l-1$, using which we derive $\hat{q}(\vF_S^L)$, required in the second and last term in \eqref{SOD_Elbo},
\begin{align}
\hat{q}(\vF_{S}^L) &= \frac{p(\vy_S|\vF_S^L)q(\vF_S^L)}{\mathcal{Z}}
= \mathcal{N}(\hat{\boldsymbol{\mu}_S^L},\hat{\boldsymbol{\Sigma}_S^L})\\
\textrm{where } \hat{\boldsymbol{\Sigma}_S^L} &= ((\boldsymbol{\Sigma}_S^L)^{-1} + (\sigma^2\boldsymbol{I})^{-1})^{-1} \nonumber \\
\hat{\boldsymbol{\mu}_S^L} &= \hat{\boldsymbol{\Sigma}_S^L}((\sigma^2\vI)^{-1}\vy_S+(\boldsymbol{\Sigma}_S^L)^{-1}\boldsymbol{\mu}_S^L) \nonumber 
\end{align}
as well as $q(\vZ_S^{l-1})$ required in the last two terms in \eqref{SOD_Elbo},
\begin{align}
q(\vZ_S^{l-1}) &= \prod_{d=1}^{D^{l-1}}\int p(\vZ_{S,d}^{l-1}|\vF_{S,d}^{l-1})q(\vF_{S,d}^{l-1})d\vF_{S,d}^{l-1}\\
& = \prod_{d=1}^{D^{l-1}} \mathcal{N}(\vZ_{S,d}^{l-1}|\boldsymbol{\mu}_{S,d}^{l-1},\sigma^2 \pmb{I} + \boldsymbol{\Sigma}_{S,d}^{l-1}) .
\end{align}

We also need the marginal distribution $q(\vF_{\bar{S}}^L)$ to compute the expectation in the first term in \eqref{SOD_Elbo}. This can be obtained by integrating out all the latent variables except $\vF_{\bar{S}}^L$ in the variational posterior approximation \eqref{qPosterior}. The latent function values in the intermediate layers $\vF^l$ can be integrated out and consequently the  marginal can be written as~\footnote{A detailed derivation of the lower bound and $q(\vF_{\bar{S}}^L)$ is provided in the supplementary material. } 
\begin{align}
\int q(\vF_{\bar{S}}^L|\vZ^{L-1},\vy_S)\left(\prod_{l=2}^{L-1}q(\vZ^{l}|\vZ^{l-1})\right)q(\vZ^1|\vX) d\vZ^{1:L-1}. \label{marginalFsbar}
\end{align}
where the first term is a Gaussian: $q(\vF_{\bar{S}}^L|\vZ^{L-1},\vy_S) = \int \hat{q}(\vF_S^L)p(\vF_{\bar{S}}^L|\vF_S^L,\vZ^{L-1}) d\vF_S^L$.  The conditional probability $q(\vZ^{l}|\vZ^{l-1})$  are obtained by integrating out $\vF^l$ and also follows a Gaussian distribution. However, the latent variable $\vZ^{l}$ depend non-linearly on $\vZ^{l-1}$ through the kernel and hence the  marginal  can not  be  computed  analytically. We use Monte-Carlo sampling to obtain the samples from the marginal and recursively draw the $i$'th sample $\hat{\vZ}^l_{(i)} \sim q(\vZ^l|\vZ^{l-1}_{(i)})$ for $l=1,2,\ldots,L-1$ with $\hat{\vZ}^0=\vX$ and use in $q(\vF_{\bar{S}}^L|\vZ^{L-1},\vy_S)$ to obtain $\vF_{\bar{S}}^L$ samples.  
\begin{equation}
q({\vF_{\bar{S}}^{L}}) = \frac{1}{T}\sum_{i=1}^{T} q(\vF_{\bar{S}}^L|\hat{\vZ}^{L-1}_{(i)},\vy_S)
\end{equation}
In order to facilitate gradient computation, we use the re-parameterization trick to obtain the samples. 

The variational parameters  $\vlambda_S = \{{\boldsymbol{\mu}}_{S}^l,{\boldsymbol{\Sigma}}_{S}^l\}_{l=1}^L$   and kernel hyper-parameters are learnt by maximizing the   variational lower bound  \eqref{SOD_Elbo}.  Computation of proposed lower bound has complexity of $\mathcal{O}(NM^2DL)$, where $M$ is the subset size, $N$ is number of data points,  $L$ is number of layers and $D = \max\{D^1,D^2,\ldots,D^L\}$

Prediction can be performed in a similar fashion as the computation of $q(\vF^L_{\bar{S}})$, where instead of the data points not in the subset ($\vX_{\bar{S}}$),  test data points are used and we sample the final layer function values associated with the test samples using \eqref{marginalFsbar}. Predictive distribution of  $\vF^L_*$ for  test data points $\vX_{*}$   is computed using samples of $\vZ^{L-1}_*$ and $\vZ^{L-1}_S$ which are in turn obtained through the reparameterization trick.
\begin{align}
    q(\vF^L_*) = \frac{1}{T} \sum_{i=1}^{T}q(\vF^L_*|\{\hat{\vZ}_{*,(i)},\hat{\vZ}_{S,(i)}\}^{L-1},\vy_S)\label{prediction}
\end{align}

\begin{table*}[!ht]
\caption{Negative log predictive probability (NLPP) score for various DGP inference techniques on UCI regression datasets.   For each dataset, $N$ represents number of training samples and $D$ represent the input dimension.  Mean NLPP scores are reported averaged over 5 runs with variance inside the brackets.  Best performing model (lowest NLPP score) for each dataset is highlighted in the respective column. SoD-DGPx indicates our method. SGHMC-DGPx is the approach of \citet{havasi2018inference}. DSVI is the approach of \citet{dsvi}. SoD* is same as SoD  but with randomly selected subset of data points. DSVI* is same as DSVI but the inducing inputs fixed and not trained. SVGP is the approach of \citet{titsias2009variational}.}
\begin{center}
\begin{tabularx}{\textwidth}{|c|X|X|X|X|X|X|X|}
\hline
\multirow{2}{*}{\bf Model} &\multicolumn{1}{c|}{\bf Boston} &\multicolumn{1}{c|}{\bf Concrete} &\multicolumn{1}{c|}{\bf Energy} &\multicolumn{1}{c|}{\bf Winered} &\multicolumn{1}{c|}{\bf Protein} &\multicolumn{1}{c|}{\bf Naval} &\multicolumn{1}{c|}{\bf Year}  \\
&N=506 D=13 &N=1030 D=8 & N=768  D=8 &N=1599 D=22 &N=45730 D=9 &N=11934 D=26 & N=515344 D=90\\
\hline 
  &  &  &  &  & & &  \\
SoD-DGP1    &2.520(0.051)  &3.285(0.049)  &1.927(0.213)  &0.956(0.062)  &2.996(0.003) & \bf-8.15(0.19) & 3.695(0.071) \\
  &  &  &  &  & & &  \\
SoD-DGP2    &2.395(0.142)  &\bf3.058(0.087)  &0.737(0.089)  &\bf0.938(0.074)  &2.83(0.005) &-7.05(0.12) & 3.595(0.003) \\
  &  &  &  &  & & &  \\
SoD-DGP3    &\bf2.366(0.114)  &3.060(0.081)  &0.654(0.148)  &1.023(0.092)  & 2.753(0.006)  & -7.26(0.25) & 3.587(0.002)\\
  &  &  &  &  & & &  \\
SoD-DGP4    &2.430(0.185) &3.058(0.086)  &\bf0.568(0.106)  & 0.957(0.076)  &3.028(0.245)  &-6.99(0.29) & 3.582(0.005)\\
  &  &  &  &  & & &  \\
  SoD*-DGP2    & 2.606(0.022) & 3.223(0.042)  & 1.227(0.167) & 0.979(0.062)  & 2.867(0.011) & -5.96(0.79) & 3.639(0.087) \\
    &  &  &  &  & & &  \\
SoD*-DGP3    & 3.55(0.071) & 3.284(0.015)  & 1.323(0.076)  & 1.213(0.054)  & 2.807(0.008) & -6.84(0.23)  & 3.72(0.107) \\
  &  &  &  &  & & &  \\
SoD*-DGP4    &  3.55(0.071) & 3.635(0.088) & 1.799(0.159)  & 1.213(0.054)  & 2.788(0.009)  & -6.41(0.45) & 3.696(0.108) \\
  &  &  &  &  & & &  \\
SGHMC-DGP2 &3.217(0.442)  &3.484(0.292)  &3.270(5.602)  &2.696(1.828)  &2.789(0.034) &-5.49(0.82) & 3.408(0.010) \\
  &  &  &  &  & & &  \\
SGHMC-DGP3 &5.026(0.861) &3.384(0.224)  &1.636(2.342)  &3.133(1.575)  &2.782(0.062) &-5.43(0.83)  & \bf3.397(0.002)  \\
  &  &  &  &  & & &  \\
SGHMC-DGP4 &7.736(2.440) &3.856(0.574) &2.097(3.663) &2.639(2.162)  &2.743(0.025)  &-5.51(0.75)  & 3.388(0.003)\\
  &  &  &  &  & & &  \\
DSVI-DGP2   &2.43(0.062)  &3.105(0.05)  &0.761(0.119)  &0.951(0.058)  &2.815(0.010)  &-6.97(0.06)  & 3.587(0.004)\\
  &  &  &  &  & & &  \\
DSVI-DGP3   &2.427(0.059) &3.114(0.053) &0.742(0.134) &0.951(0.057) &2.755(0.004) &-6.69(0.37)  & 3.577(0.004)\\
  &  &  &  &  & & &  \\
DSVI-DGP4   &2.429(0.052) &3.127(0.066) &0.732(0.131) &0.951(0.057) &\bf2.733(0.013) &-5.07(1.88)  & 3.575(0.004)\\
  &  &  &  &  & & &  \\
DSVI*-DGP2   &2.534(0.066) &3.185(0.026)  &1.261(0.053)  &0.969(0.061)  &2.871(0.006)  &-6.30(0.25) &  3.597(0.006)\\
  &  &  &  &  & & &  \\
DSVI*-DGP3   &2.535(0.064) &3.190(0.025)  &1.270(0.061) &0.970(0.061)  &2.835(0.012)  &-5.32(1.33)  & 3.583(0.005) \\
  &  &  &  &  & & &  \\
DSVI*-DGP4   &2.537(0.063)  &3.192(0.030)  &1.296(0.036)  & 0.969(0.06)  &2.983(0.201)  &-2.80(0.02)  & 3.581(0.005) \\
  &  &  &  &  & & &  \\
 SVGP &2.455(0.054) &3.156(0.023)  &1.282(0.056)  &0.953(0.059)  &2.911(0.009)  & -7.42(0.18)  & 3.600(0.005) \\
\hline
\end{tabularx}
\end{center}
\label{nlpp_table}
\end{table*}

\section{Experiments}

We conduct experiments to evaluate the performance of the proposed inference technique for deep Gaussian processes on various regression datasets~\footnote{Code available at https://github.com/brain-iith/SOD\_DGP}.  The proposed inference technique is compared against baselines and the existing state-of-the-art inference techniques for DGPs, to demonstrate its effectiveness.

{\bf Architecture}
The DGP architecture is chosen to the same as that of~\citep{dsvi}. Input layer of the DGP has D nodes, where D is dimension of input data point. In case of regression task, final layer or output layer has number of nodes set to one.
Number of nodes are same accross all hidden or latent layers, with each latent or hidden layer has min(30,D) number of nodes. 

{\bf Subset Selection}  We use the subset of training data points as the inducing inputs. We choose the subset as the collection of data points closest to the centroids obtained after K-means clustering, where $K$ is set as the subset size ($M$). Subset size of 50 and 100 is used for small and medium size datasets respectively.

{\bf Model and Variational Parameters} For variational distribution $q(\vF^l_{S})$, all variational mean vectors are initialised with random vectors and variational covariance matrix are initialised with identity matrix(scaled by $10^{-5}$ except the final layer). 
The hyper-parameters associated with the model are the kernel parameters and noise variance in each layer. 
For all the reported results for our model, both kernel variance and lengthscale parameters are initialised with the value of 0.5, and noise variance is initialised to 0.01 in the final layer and $10^{-5}$ in the intermediate layers.

\begin{figure*}[t]
\hspace*{-5mm}
     \centering
     \begin{subfigure}
      \centering
      \includegraphics[width=17.5cm]{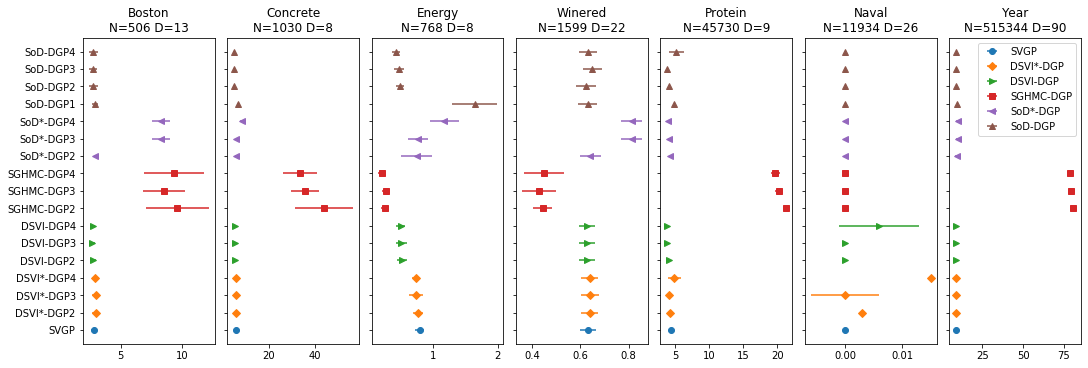}
    \end{subfigure}
\caption{ RMSE scores (mean along horizontal bars representing the standard deviation) for various DGP inference techniques on the UCI Regression datasets. Lower values (to the left) provide  better results. }     
\label{test_rmse}
\end{figure*}

{\bf Evalauation Metrics} Negative log predictive probability (NLPP) or negative log likelihood and root mean square error (RMSE) on test data  is used to report performance on regression datasets. NLPP score consider  the confidence in the predictions (lower is better) and are more significant in evaluating performance of probabilistic models. While, RMSE computes the error between point predictions (mean values) and the actual observed outputs (lower is better) ignoring the variance around predictions. 

{\bf Training and Preprocessing}
Inputs and outputs in training set are scaled to zero mean unit standard deviation and the same scaling is applied in the test data set for evaluation. We consider multiple random splits of training and test data with $10\%$ of  the data as test data.  We report the  average  RMSE and NLPP scores averaged over 5 runs.  Training (lower bound maximization) is done for 20,000 iterations with batch size of 2000, using Adam optimizer initialised with learning rate of 0.01. For sampling the latent representations, we use 10 samples during training and 50 samples during prediction.  

{\bf Baselines }
 The DGP with the proposed subset of data (SoD) inference technique (SoD-DGP) is compared against a  single layer sparse variational GP (SVGP)~\citep{titsias2009variational}, and  
 state-of-the-art DGP approaches  such as doubly stochastic variational inference based DGP (DSVI-DGP)~\citep{dsvi} and stochastic gradient Hamiltonian Monte Carlo (SGHMC) based DGP (SGHMC-DGP)~\citep{havasi2018inference}. All the previous approaches consider inducing inputs in each layer as learnable parameters  while doing  approximate inference for DGPs.  The proposed approach use a subset of dataset  as  inducing inputs and in each layer the inducing inputs are obtained by sampling from the conditional distribution $q(
 \vZ^{l}|\vZ^{l-1})$. We also consider a baseline, SoD*-DGP  to check the sensitivity of the subset selection strategy towards the SoD-DGP performance. SoD*-DGP uses the same  SoD-DGP variational lower bound to learn the parameters but subset selection is done by randomly choosing the subset of data points instead of K-Means clustering based selection.  Similarly, we consider a baseline DSVI*-DGP,  where inference is done using DSVI but inducing inputs are not treated as learnable parameters but are fixed to the initial values.  Here, inducing inputs for first layer is initialized based on centroids of K-means clustering, and then a  PCA mapping  is done as discussed in \cite{dsvi} to initialize intermediate layer inducing inputs. Unlike DSVI*-DGP, inducing input samples in  SoD-DGP  changes after every iteration as the conditional distribution $q(\vZ^{l}|\vZ^{l-1})$ evolves. The advantage of the proposed approach is visible from the experimental results.
Model hyperparameters for the baseline methods are initialised and tuned with the values reported for these approaches~\citep{dsvi,havasi2018inference}.

{\bf Regression Results }
We consider 7 standard small to large sized UCI regression benchmark datasets: Boston, Concrete, Energy, Winered, Protein, Naval and Year to evaluate the performance of the models. Table~\ref{nlpp_table} and Fig.~\ref{test_rmse} provide the NLPP and RMSE results (mean and standard deviation) respectively for  DSVI-DGP, DSVI*-DGP, SGHMC-DGP and the proposed approach SoD-DGP~\footnote{Exact values vary slightly from the prior work because the training and test splits used are different.}.  We consider DGP models with different number of hidden layers  DGP1 (1 hidden layer), DGP2 (2 hidden layers), DGP3 (3 hidden layers) and DGP4 (4 hidden layers). With respect to NLPP score, we can observe from Table~\ref{nlpp_table} that the SoD-DGP approach performs better than other DGP inference approaches in all the datasets except Protein where DSVI-DGP gives the best performance. In this case also, SoD-DGP performance is very close. We find that the DSVI-DGP models perform better than DSVI*-DGP models, which does not optimize and learn the inducing inputs but fixes it to initial values. We can also observe that SoD-DGP results are in general better than SoD*-DGP, which considers random subset of data points. We find this to be more observable in deeper DGP models, while SoD*-DGP2 can provide results closer to SoD-DGP models. Thus, SoD-DGP models can be sensitive to the subset used and their performance can be further improved using a better subset selection strategy than the presently used naive K-Means clustering subset selection. The SVGP model gives a competitive performance on relatively `easy' regression datasets as reported also in \citet{dsvi}. In the easy `Naval' dataset, all the methods show very low NLPP values~\footnote{For Naval, NLPP is negative and RMSE is close to zero as it is a simpler problem.}  and SVGP gives best performance while SoD-DGP gives second best result.  The NLPP results also suggest that SoD-DGP provides better confidence in its predictions and might have better uncertainty modelling capabilities. 

Fig.~\ref{test_rmse} provides RMSE results for various inference techniques for DGP models with lower values towards left side giving better results. We can observe that SoD-DGP models performs better or close to DSVI-DGP models and gives best performance for most of the data sets. In 2 datasets, Energy and Winered,  SGHMC-DGP gives better performance. Similar to NLPP results, SoD-DGP and DSVI-DGP performs better than DSVI*-DGP. Thus, the proposed SoD-DGP model provides a computational advantage in terms of reducing the number of parameters while  improving or maintaining the generalization performance.

\section{Conclusion}
We have proposed a new inference technique for deep Gaussian processes which could overcome some limitations of the existing inference techniques for DGPs. Existing inference technique require estimating large number of inducing inputs which grows with the number of layers. We propose an inference technique where inducing inputs are set to a fixed subset in training data for the first layer and the inducing inputs for subsequent layers are sampled from the conditional variational posterior. This approach reduces the number of parameters to be estimated while maintaining the generalization performance of the DGP model. This is evident from the experimental results on UCI regression datasets,  where we found that SoD-DGP gives better log-likelihood values than other DGP inference techniques. The proposed approach is also amenable to sampling techniques like leverage scoring and determinantal point processes which could further improve the performance through better choice of subsets. As a future work, we will extend the proposed approach for classification problems.  

\textbf{Acknowledgement:} MEK would like to thank a number of people who, over the years, spent time in checking the validity of \eqref{sod_vlb}, including Wu Lin, Heiko Strathman, Didrik Nielsen, Si Kai Lee, Anand Subramanian, Paul Cheng, and Arno Solin. AJ and PKS thank the funding and travel support from Science and Engineering Research Board (SERB), India and Japan International Co-operation Agency (JICA), Japan.

\bibliography{jain_527}

\onecolumn

\section*{Appendix}

\section*{A. Training and Inference in Deep Gaussian Process Models}

The sparse variational inference approach simultaneously addresses intractability and scalability issues in Deep GPs. This introduces variational parameters and inducing points with inducing input $\bar{\vZ}^l$ and inducing outputs $\vU^l$ for each layer $l$, all of which are learnt from the variational lower bound.
\begin{figure}[t]
  \centering
  \includegraphics[scale=0.5]{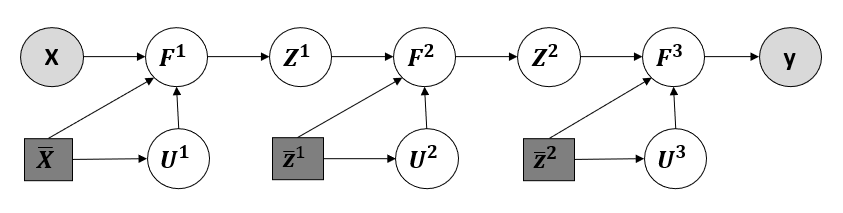}
  \caption{Deep Gaussian Process Model with inducing points as in ~\cite{damianou13}}
  \label{andreasAugDGP}
\end{figure}
\begin{figure}[t]
  \centering
  \includegraphics[scale=0.5]{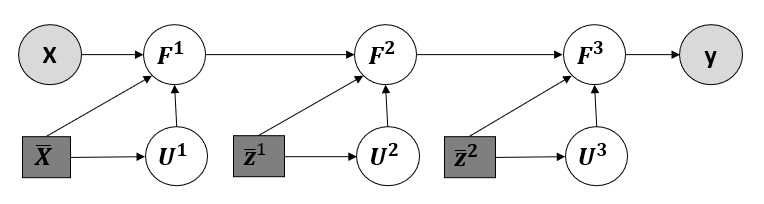}
  \caption{Deep Gaussian Process Model with inducing points in ~\cite{dsvi}}
  \label{dsviAugDGP}
\end{figure}
\begin{figure}[t]
  \centering
  \includegraphics[scale=0.5]{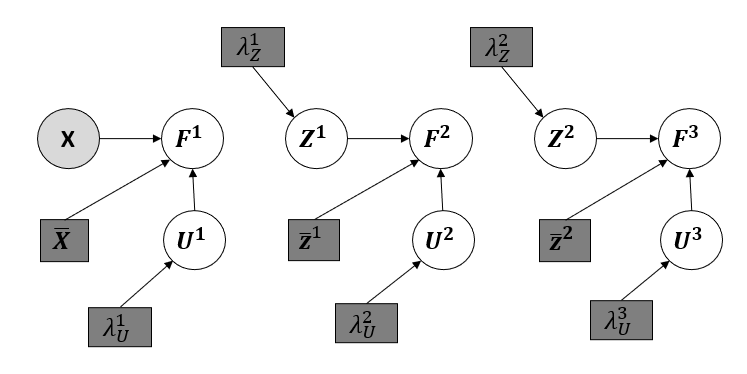}
  \caption{Approximate Inference For Deep Gaussian Process in ~\cite{damianou13}}
  \label{andreasDGP}
\end{figure}
\begin{figure}[ht]
  \centering
  \includegraphics[scale=0.5]{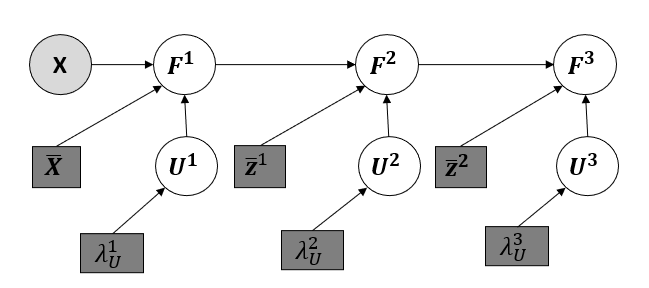}
  \caption{Doubly Stochastic Variational Inference For Deep Gaussian Process ~\cite{dsvi}}
  \label{DSVIDGP}
\end{figure}

\begin{figure}[t]
  \centering
  \includegraphics[scale=0.5]{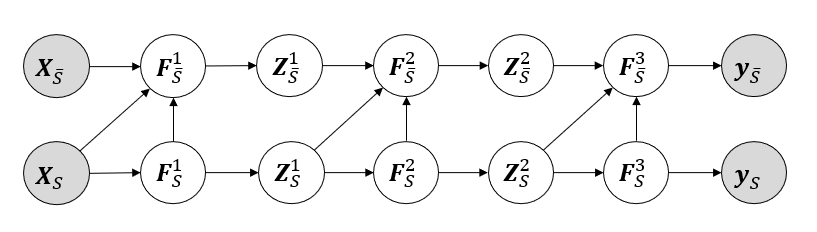}
  \caption{Proposed Subset-of-Data Representation of Original Deep Gaussian Process}
  \label{partitionedDGP}
\end{figure}
\begin{figure}[t]
  \centering
  \includegraphics[scale=0.5]{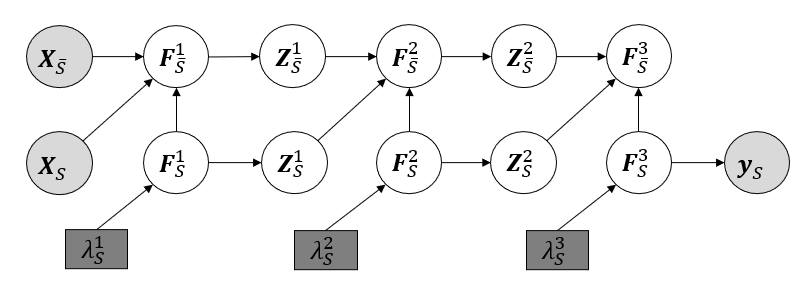}
  \caption{Proposed Subset-of-Data Variational Inference For Deep Gaussian Process}
  \label{SODVIDGP}
\end{figure}
Figure~\ref{andreasAugDGP}  and  \eqref{damianouDGPLik} provides the graphical model and full likelihood respectively for sparse Deep GP model introduced in ~\cite{damianou13}. While, Fig.~\ref{dsviAugDGP} and \eqref{dsviDGPLik} provides them for the sparse deep GP introduced in \cite{dsvi}.  The later consider the noise inside the kernel and hence drops the explicit noisy representations $\vZ^l$ from the model. 
\begin{align}
 p({\vy}|\vF^{L})\prod_{l=L-1}^{1} p(\vF^{l+1}|\vU^{l+1},\vZ^l,\bar{\vZ}^l)p(\vU^{l+1}|\bar{\vZ}^l)\times
 p(\vZ^l|\vF^l)p(\vF^1|\vX,\bar{\vX},\vU^1)p(\vU^1|\bar{\vX}) \label{damianouDGPLik}\\
 p({\vy}|\vF^{L})\prod_{l=L-1}^{1} p(\vF^{l+1}|\vU^{l+1},\vF^l,\bar{\vZ}^l)p(\vU^{l+1}|\bar{\vZ}^l)\times
 p(\vF^1|\vX,\bar{\vX},\vU^1)p(\vU^1|\bar{\vX}) \label{dsviDGPLik}
\end{align}
Here, $\bar{\vZ}^l$ are pseudo inducing inputs at layer l, $\bar{\vX}$ pseudo inputs at layer 1, $\vU^l = f(\bar{\vZ}^l)$ are inducing points which are function values evaluated at corresponding pseudo inputs. For simplicity we will avoid the use of $\bar{\vZ}^l$ and $\bar{\vX}$ in further discussions.\\

The marginal likelihood and posterior computation are intractable in these models because $\vZ^l$ and $\vF^l$ are appearing in non linear manner inside the covariance matrices of distributions $p(\vF^{l+1}|\vU^l,\vZ^l)$ and $p(\vF^{l+1}|\vU^l,\vF^l)$  in  \eqref{damianouDGPLik} and \eqref{dsviDGPLik} respectively. Both \cite{damianou13} and \cite{dsvi} obtain tractable lower bound to marginal likelihood using variational inference by introducing an approximate variational posterior distribution. The deep GP inference in \cite{damianou13} used a mean field variational approximation while \cite{dsvi} used a doubly stochastic variational inference which maintained the conditional structure of deep GP.  Equation \eqref{damianouVarPost} shows the variational posterior used in ~\cite{damianou13} where $q(\vU^l)$ and $q(\vZ^l)$ are variational distributions factorised accross the layers.
Variational distribution over $\vZ^l$ is independent of the input from previous layer which  disconnect the link between the layers and result in losing the correlations between layers as shown in Fig.~\ref{andreasDGP}.
\begin{align}
q(\vF,\vU,\vZ) = \prod_{l=L-1}^{1}p(\vF^{l+1}|\vU^{l+1},\vZ^l)q(\vU^{l+1})\times\nonumber\\
q(\vZ^l)p(\vF^1|\vU^1,\vX)q(\vU^1) \label{damianouVarPost}
\end{align}
The variational posterior used in ~\cite{dsvi} is shown in \eqref{dsviDGPVarPost}. Here, $q(\vU^l)$ is a variational distribution factorised across the layers while $q(\vF^l)$ ( $q(\vF^l) = \bar{q}(\vF^l|\vF^{l-1}) = \int p(\vF^l|\vU^l,\vF^{l-1}) q(\vU^l) d\vU^l$) maintains the dependence between the layers and full conditional structure as shown in Fig.~\ref{DSVIDGP}.  
\begin{align}
q(\vF,\vU) = \prod_{l=L-1}^{1}p(\vF^{l+1}|\vU^{l+1},\vF^l)q(\vU^{l+1})\times\nonumber\\
p(\vF^1|\vU^1,\vX)q(\vU^1) \label{dsviDGPVarPost}
\end{align}
\section*{B. Proposed Inference Method for Deep Gaussian Processes}
\subsection*{B.1 Evidence Lower Bound}
We learn the variational parameters and hyper-parameters by maximizing the evidence lower bound.  We derive the evidence lower bound (ELBO) for the  proposed approach  as follows. 
\begin{align*}
ELBO &= E_{q\left(\vF^{L},\left\{\vF^{l},\vZ^{l}\right\}_{l=1}^{L-1}\right)}\left[\log\left(\frac{p(\vy,\left\{\vF^l,\vZ^{l-1}\right\}_{l=1}^{L})}{q\left(\vF^{L},\left\{\vF^{l},\vZ^{l}\right\}_{l=1}^{L-1}\right)}\right)\right] \\
\text{where } p(\vy,\left\{\vF^l,\vZ^{l-1}\right\}_{l=1}^{L}) &= \prod_{i=1}^{N}p(y_i|f_i^L) \left(\prod_{l=2}^Lp(\vF^l|\vZ^{l-1})p(\vZ^{l-1}|\vF^{l-1})\right)p(\vF^1|\vX)
\end{align*}
\begin{align*}
\text{and } q\left(\vF^{L},\left\{\vF^{l},\vZ^{l}\right\}_{l=1}^{L-1}\right)
&= p(\vy_S|\vF_S^L)\left(\prod_{l=2}^{L}p(\vF_{\bar{S}}^{l}|\vF_S^l,\vZ^{l-1})q(\vF_S^l)p(\vZ^{l-1}|\vF^{l-1})\right)p(\vF_{\bar{S}}^{1}|\vF_S^{1},X)q(\vF_S^{1})\frac{1}{\mathcal{Z}}\label{qPosterior}\\
&= \prod_{n \in S}p(y_n|f_n^L) \left(\prod_{l=2}^{L}\prod_{d=1}^{D^{l}}p(\vf_{\bar{S},d}^{l}|\vf_{S,d}^{l},\vZ^{l-1})q(\vf_{S,d}^{l})\prod_{n =1}^{N}\mathcal{N}({ Z_{n,:}^{l-1} };{\vf_{n,:}^{l-1} }, \sigma^2 I)\right) \nonumber\\
&\prod_{d=1}^{D^{1}}p(\vf_{\bar{S},d}^{1}|\vf_{S,d}^{1},\vX)q(\vf_{S,d}^{1})\frac{1}{\mathcal{Z}} 
\end{align*}
In the ELBO derivation, a few terms inside the logarithm cancel due to the factorisation assumed in the variational distribution
\begin{align*}
ELBO &= E_{q}\left[\log\left( \frac{\prod_{i=1}^{N}p(y_i|\vf_i^L) \left(\prod_{l=2}^Lp(\vF^l|\vZ^{l-1})p(\vZ^{l-1}|\vF^{l-1})\right)p(\vF^1|\vX)\mathcal{Z}}{p(\vy_S|\vF_S^L)\left(\prod_{l=2}^{L}p(\vF_{\bar{S}}^{l}|\vF_S^l,\vZ^{l-1})q(\vF_S^l)p(\vZ^{l-1}|\vF^{l-1})\right)p(\vF_{\bar{S}}^{1}|\vF_S^{1},X)q(\vF_S^{1})}\right) \right]\\
 &= E_{q\left(\vF^{L},\left\{\vF^{l}\vZ^{l}\right\}_{l=1}^{L-1}\right)}\left[\log\left(\frac{p(\vy_{\bar{S}}|\vF_{\bar{S}}^L) \left(\prod_{l=2}^{L}p(\vF_S^l|\vZ_S^{l-1})\right)p(\vF_S^1|\vX_S) \mathcal{Z}}{\prod_{l=1}^L q(\vF_S^l)}\right)\right]\\
&= E_{q(\vF_{\bar{S}}^L)}[log(p(\vy_{\bar{S}}|\vF_{\bar{S}}^L))] + log(\mathcal{Z}) - \sum_{l=2}^{L-1} \sum_{d=1}^{D^l} E_q(\vZ_S^{l-1})\left[ KL(q(\vf_{S,d}^l)||p(\vf_{S,d}^l|\vZ_S^{l-1}))\right] \nonumber \\
& - \sum_{d=1}^{D^1} KL(q(\vf_{S,d}^1)||p(\vf_{S.d}^1|\vX_S)) +  E_{q(\vZ_S^{L-1})}\left[E_{\hat{q}(\vF_{S}^L)}\left[log\left(\frac{p(\vF_{S}^L|\vZ_S^{L-1})}{q(\vF_{S}^L)}\right)\right]\right] \\
&= E_{q(\vF_{\bar{S}}^L)}[log(p(\vy_{\bar{S}}|\vF_{\bar{S}}^L))] + log(\mathcal{Z}) - \sum_{l=2}^{L-1} \sum_{d=1}^{D^l} E_q(\vZ_S^{l-1})\left[ KL(q(\vf_{S,d}^l)||p(\vf_{S,d}^l|\vZ_S^{l-1}))\right] \nonumber \\
& - \sum_{d=1}^{D^1} KL(q(\vf_{S,d}^1)||p(\vf_{S.d}^1|\vX_S)) +  E_{q(\vZ_S^{L-1})}\left[E_{\hat{q}(\vF_{S}^L)}\left[log\left(\frac{p(\vF_{S}^L|\vZ_S^{L-1})\hat{q}(\vF_S^L)}{\hat{q}(\vF_S^L)q(\vF_{S}^L)}\right)\right]\right] \\
\end{align*}
\begin{align*}
& = E_{q(\vF_{\bar{S}}^L)}[\log(p(\vy_{\bar{S}}|\vF_{\bar{S}}^L))] + \log(\mathcal{Z}) - \sum_{l=2}^{L-1} \sum_{d=1}^{D^l} E_{q(\vZ_S^{l-1})}\left[ KL(q(\vf_{S,d}^l) \|p(\vf_{S,d}^l | \vZ_S^{l-1}))\right]\nonumber\\
& - \sum_{d=1}^{D^1} KL(q(\vf_{S,d}^1) \|p(\vf_{S.d}^1|\vX_S)) -  E_{q(\vZ_S^{L-1})}[KL(\hat{q}(\vF_{S}^L)\|p(\vF_{S}^L |\vZ_S^{L-1}))] \nonumber\\
&+KL(\hat{q}(\vF_{S}^L)\|q(\vF_{S}^L)) \\
& = E_{q(\vF_{\bar{S}}^L)}[\log(p(\vy_{\bar{S}}|\vF_{\bar{S}}^L))] + \log(\mathcal{Z}) - \sum_{l=2}^{L-1} \sum_{d=1}^{D^l} E_{q(\vZ_S^{l-1})}\left[ KL(q(\vf_{S,d}^l) \|p(\vf_{S,d}^l | \vZ_S^{l-1}))\right]\nonumber\\
& - \sum_{d=1}^{D^1} KL(q(\vf_{S,d}^1) \|p(\vf_{S.d}^1|\vX_S)) -  E_{q(\vZ_S^{L-1})}[KL(\hat{q}(\vF_{S}^L)\|p(\vF_{S}^L |\vZ_S^{L-1}))] \nonumber\\
&+E_{\hat{q}(\vF_{S}^L)}[log(p(\vy_S|\vF_{S}^L))]- \log(\mathcal{Z})\\
& = E_{q(\vF_{\bar{S}}^L)}[\log(p(\vy_{\bar{S}}|\vF_{\bar{S}}^L))] +E_{\hat{q}(\vF_{S}^L)}[log(p(\vy_S|\vF_{S}^L))] - \sum_{d=1}^{D^1} KL(q(\vf_{S,d}^1) \|p(\vf_{S.d}^1|\vX_S))\\
&- \sum_{l=2}^{L-1} \sum_{d=1}^{D^l} E_{q(\vZ_S^{l-1})}\left[ KL(q(\vf_{S,d}^l) \|p(\vf_{S,d}^l | \vZ_S^{l-1}))\right]\nonumber  -  E_{q(\vZ_S^{L-1})}[KL(\hat{q}(\vF_{S}^L)\|p(\vF_{S}^L |\vZ_S^{L-1}))]
\end{align*}
where 
\begin{align*}
p(\vf_{\bar{S},d}^l|\vf_{S,d}^l,\vZ^{l-1}) &= \mathcal{N}({\vf^{l}_{\bar{S},d} };\vm_{\bar{S},d}^l,\vV_{\bar{S}}^l) \quad ; \quad 
\vm_{\bar{S},d}^l = \vK^{l}_{\bar{S},S}{\vK^{l}_{S,S}}^{-1}\vf_{S,d}^l \quad ; \quad 
\vV_{\bar{S}}^l = \vK^{l}_{\bar{S},\bar{S}}-\vK^{l}_{\bar{S},S}{\vK^{l}_{S,S}}^{-1}\vK^{l}_{S,\bar{S}}\\
q(\vf_{S,d}^l) &=  \mathcal{N}(\vf_{S,d}^l;{\bf{\mu}}_{S,d}^l,{\bf{\Sigma}}_{S,d}^l) \quad ; \quad
p(\vy_S|\vF_S^L) = \mathcal{N}(\vy_S;\vF_S^L,\sigma^2\vI)\\
\mathcal{Z} &= p(\mathcal{D_S})  = \int p(\vy_S|\vF_S^L)q(\vF_S^L)d\vF_S^L 
= \mathcal{N}(\vy_S|{\bf{\mu}}_S^L, \sigma^2\vI + {\bf{\Sigma}}_S^L)
\end{align*}
\subsection*{B.2 Computation of Marginal Distribution $q(\vF_{\bar{S}}^L)$}
\begin{align*}
q\left(\vF^{L},\left\{\vF^{l},\vZ^{l}\right\}_{l=1}^{L-1}\right)
&= p(\vy_S|\vF_S^L)\left(\prod_{l=2}^{L}p(\vF_{\bar{S}}^{l}|\vF_S^l,\vZ^{l-1})q(\vF_S^l)p(\vZ^{l-1}|\vF^{l-1})\right)p(\vF_{\bar{S}}^{1}|\vF_S^{1},X)q(\vF_S^{1})\frac{1}{\mathcal{Z}}
\end{align*}
\begin{align*}
\mathcal{Z} * q(\vF_{\bar{S}}^L)&= \int p(\vy_S|\vF_S^L)\left(\prod_{l=2}^{L}p(\vF_{\bar{s}}^{l}|\vF_S^l,\vZ^{l-1})q(\vF_S^l)p(\vZ^{l-1}|\vF^{l-1})\right)p(\vF_{\bar{S}}^1|\vF_S^1,\vX)q(\vF_S^1)\\
&   d\vF_S^Ld\left\{\vF^{l},\vZ^{l}\right\}_{l=1}^{L-1}
\end{align*}
\begin{align*}
\mathcal{Z} * q(\vF_{\bar{S}}^L)& = \int (p(\vy_S|\vF_S^L)q(\vF_S^L))p(\vF_{\bar{S}}^L|\vF_S^L,\vZ^{L-1})d\vF_S^L \left(\prod_{l=2}^{L-1}q(\vZ^{l}|\vZ^{l-1}\right)q(\vZ^1|\vX)d\left\{\vZ^{l}\right\}_{l=1}^{L-1}\\
\end{align*}
Using Bayes theorem to rewrite first term in integral
\begin{align*}
\bcancel{\mathcal{Z}} * q(\vF_{\bar{S}}^L)& = \int (\bcancel{\mathcal{Z}}q(\vF_S^L|\vy_S))p(\vF_{\bar{S}}^L|\vF_S^L,\vZ^{L-1})d\vF_S^L \left(\prod_{l=2}^{L-1}q(\vZ^{l}|\vZ^{l-1}\right)q(\vZ^1|\vX)d\left\{\vZ^{l}\right\}_{l=1}^{L-1}\\
q(\vF_{\bar{S}}^L)&= \int(q(\vF_S^L|\vy_S)p(\vF_{\bar{S}}^L|\vF_S^L,\vZ^{L-1}))d\vF_S^L\left(\prod_{l=2}^{L-1}q(\vZ^{l}|\vZ^{l-1}\right)q(\vZ^1|\vX)d\left\{\vZ^{l}\right\}_{l=1}^{L-1}\\
&= \int q(\vF_{\bar{S}}^L|\vZ^{L-1},\vy_S)\left(\prod_{l=2}^{L-1}q(\vZ^{l}|\vZ^{l-1}\right)q(\vZ^1|\vX)d\left\{\vZ^{l}\right\}_{l=1}^{L-1}
\end{align*}
where,
\begin{align*}
q(\vF_{\bar{S}}^L|\vZ^{L-1},\vy_S) &= \int q(\vF_S^L|\vy_S)p(\vF_{\bar{S}}^L|\vF_S^L,\vZ^{L-1}) d\vF_S^L\\
&= \mathcal{N}(\vK^L_{\bar{S},S}(\vK^L_{S,S})^{-1}\hat{\mu_S^L},\vV_{\bar{S}}^L + \vK^L_{\bar{S},S}(\vK^L_{S,S})^{-1}\hat{\Sigma_S^L}(\vK^L_{\bar{S},S}(\vK^L_{S,S})^{-1})^T)
\end{align*}
for $2<=l<=L$
\begin{align*}
q(\vZ^{l-1}|\vZ^{l-2}) &= \int p(\vZ^{l-1}|\vF^{l-1})p(\vF_{\bar{S}}^{l-1}|\vF_S^{l-1},\vZ^{l-2})q(\vF_S^{l-1})d\vF^{l-1}\\
&= \prod_{d=1}^{D^{l-1}}\int p(Z_{:,d}^{l-1}|\vf_{:,d}^{l-1})p(\vf_{\bar{S},d}^{l-1}|\vf_{S,d}^{l-1},\vZ^{l-2})q(\vf_{S,d}^{l-1})d\vf_{:,d}^{l-1}\\
& =\prod_{d=1}^{D^{l-1}}\int p(Z_{:,d}^{l-1}|\vf_{:,d}^{l-1})q(\vf_{:,d}^{l-1}|\vZ^{l-2})d\vf_{:,d}^{l-1}\\
&=\prod_{d=1}^{D^{l-1}}\int \mathcal{N}(Z_{:,d}^{l-1}|\vf_{:,d}^{l-1},\sigma^2\vI) \mathcal{N}(\vf_{:,d}^{l-1}|\mu_{:,d}^{l-1},\Sigma_{:,d}^{l-1})d\vf_{:,d}^{l-1}\\
&=\prod_{d=1}^{D^{l-1}}\mathcal{N}(Z_{:,d}^{l-1}|\mu_{:,d}^{l-1},\sigma^2\vI + \Sigma_{:,d}^{l-1})
\end{align*}
where,
\begin{align*}
\mu_{:,d}^{l-1} &= \begin{bmatrix}
											\mu_{S,d}^{l-1}\\
                                            \vK_{\bar{S},S}^{l-1}(\vK_{S,S}^{l-1})^{-1}\mu_{S,d}^{l-1}\\
                                            \end{bmatrix}. \quad ; \quad
\Sigma_{:,d}^{l-1} = \begin{bmatrix} A & B\\ C & D\\ \end{bmatrix}\\   
A &= \Sigma_{S,d}^{l-1} \quad ; \quad 
B = \Sigma_{S,d}^{l-1} (\vK_{S,S}^{l-1})^{-1}\vK_{S,\bar{S}}^{l-1} \quad ; \quad 
C = (\Sigma_{S,d}^{l-1} (\vK_{S,S}^{l-1})^{-1}\vK_{S,\bar{S}}^{l-1})^T  \\
D &= \vV_{\bar{S}}^{l-1} + \vK_{\bar{S},S}^{l-1}(\vK_{S,S}^{l-1})^{-1}\Sigma_{S,d}^{l-1} (\vK_{S,S}^{l-1})^{-1}\vK_{S,\bar{S}}^{l-1}\nonumber
\end{align*}

\section*{C. Experimental Results}

\begin{figure}[h]
    \centering
    \subfigure[ELBO ]{\includegraphics[width=5.2cm]{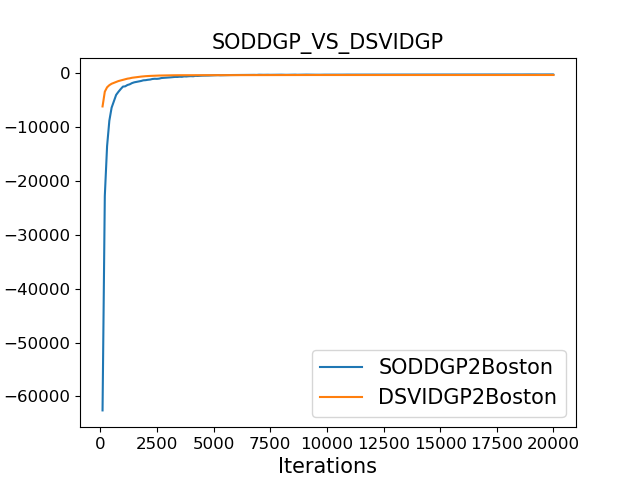}}
        \subfigure[Test Log-likeihood ]{\includegraphics[width=5.2cm]{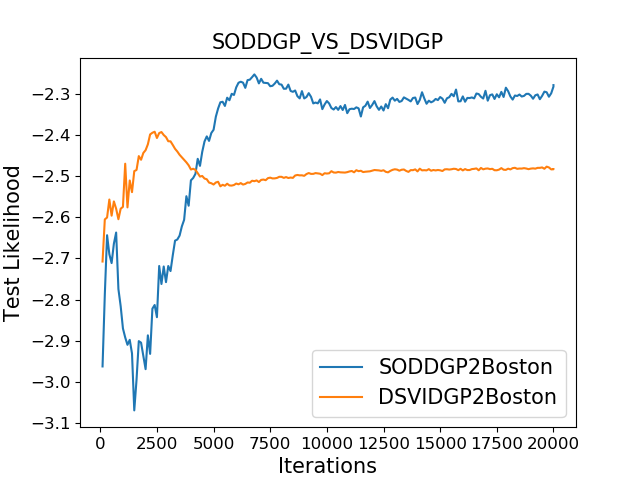}}
                \subfigure[Test RMSE ]{\includegraphics[width=5.2cm]{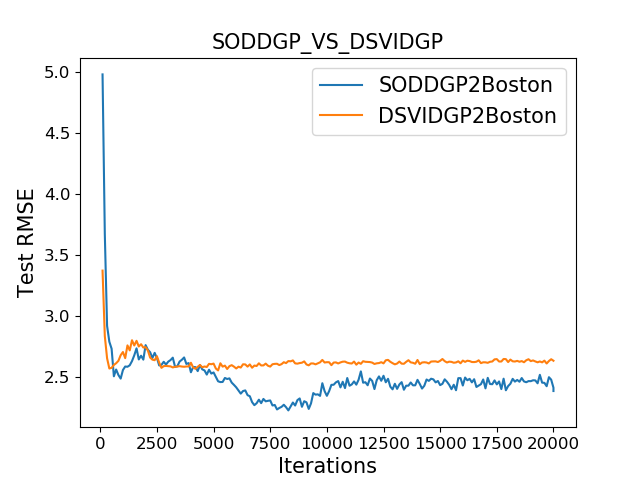}}
    \caption{ELBO, test log-likelihood and test RMSE on Boston housing}
    \label{fig:boston}
\end{figure}

We show the change in objective function (ELBO) on the training split, and log-likelihood and root mean square error on the test split for the Boston data in Figure~\ref{fig:boston} over the training iterations.  We can clearly see that the proposed SoD-DGP achieves a better performance than DSVI-DGP. 

\end{document}